\documentclass{CVIS}

\usepackage{amsmath,bm}

\usepackage{times}
\usepackage{epsfig}
\usepackage{graphicx}
\usepackage{amsmath}
\usepackage{amssymb}
\usepackage{soul,color}
\usepackage{epstopdf}

\begin{document}
	
	\title{Discovery Radiomics via Deep Multi-Column Radiomic Sequencers for Skin Cancer Detection}
	
	\author{
		\begin{tabularx}{\textwidth}{X X}
			Mohammad Javad Shafiee & University of Waterloo, ON, Canada  \\
			&Elucid Labs, Canada \\
			Alexander Wong& University of Waterloo,  ON, Canada  \\
			&Elucid Labs, Canada \\		
		\end{tabularx}
	}
	
	\maketitle

%%%%%%%%% ABSTRACT
\begin{abstract}
While skin cancer is the most diagnosed form of cancer in men and women, with more cases
diagnosed each year than all other cancers combined, sufficiently early diagnosis results
in very good prognosis and as such makes early detection crucial.  While radiomics
 have shown considerable promise as a powerful diagnostic tool for significantly improving
 oncological diagnostic accuracy and efficiency, current radiomics-driven methods have largely
  rely on pre-defined, hand-crafted quantitative features, which can greatly limit the ability to
  fully characterize unique cancer phenotype that distinguish it from healthy tissue.  Recently,
  the notion of discovery radiomics was introduced, where a large amount of custom, quantitative radiomic features
  are directly discovered from the wealth of readily available medical imaging data. In this study, we present a novel
  discovery radiomics framework for skin cancer detection, where we leverage novel deep multi-column
  radiomic sequencers for high-throughput discovery and extraction of a large amount of
  custom radiomic features tailored for characterizing unique skin cancer tissue phenotype.
  The discovered radiomic sequencer was tested against 9,152 biopsy-proven clinical images comprising of different skin
  cancers such as melanoma and basal cell carcinoma, and demonstrated
  sensitivity and specificity of 91\% and 75\%, respectively, thus achieving dermatologist-level
  performance and \break hence can be a powerful tool for assisting general practitioners and dermatologists
  alike in improving the efficiency, consistency, and accuracy of skin cancer diagnosis.

\end{abstract}

%%%%%%%%% BODY TEXT
\section{Introduction}
 \begin{figure*}
 	\begin{center}
 		\includegraphics[width = 15 cm]{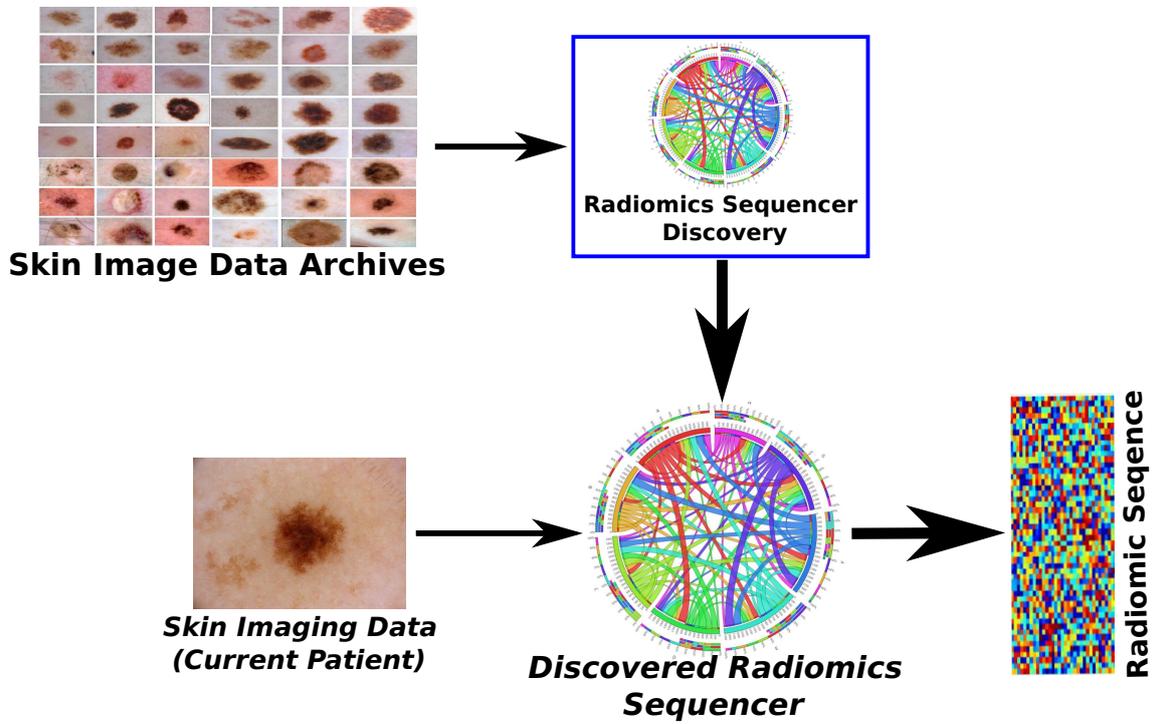}
 	\end{center}
 	\caption{Overview of the proposed discovery radiomics framework for skin cancer detection. A custom
 		radiomic sequencer is discovered via past skin imaging data; for new patients, radiomic
 		sequences of custom radiomic features are generated for skin cancer quantification and analysis. In this study, we leverage a deep multi-column radiomic sequencer to characterize and model unique skin cancer phenotype.}
 	\label{fig:skinradiomics}
 \end{figure*}

Skin cancer is the most diagnosed form of cancer, with more new cases of skin cancer
diagnosed each year than all other forms of cancers combined~\cite{Stern}.  Furthermore, the annual
cost for the treatment of skin cancer in the U.S. is estimated at \$8.1 billion~\cite{Guy}.  Fortunately,
there are high chance of  prognosis for various forms of skin cancer given sufficiently early diagnosis,
making early skin cancer screening and detection crucial for patient recovery.  A powerful
diagnostic tool that has shown considerable promise for ushering in a new era of imaging-driven quantitative personalized cancer decision support and management is the notion of radiomics~\cite{radiomics}, which involves the high-throughput extraction and analysis of a large number of quantitative features to characterize cancer tissue traits to improve oncological diagnostic efficiency, consistency, and accuracy.

Despite its considerable promise~\cite{radiomics2,radiomics3}, current radiomics-driven methods have largely relied on predefined, hand-crafted imaging-based feature models based on human notions of intensity, texture, and shape, and as such can greatly limit the ability to
  fully characterize unique cancer phenotype that distinguish it from healthy tissue.  Recently, to alleviate the limitations of radiomics, the notion of discovery radiomics was introduced~\cite{discoveryradiomics2,discoveryradiomics3,discoveryradiomics1,discoveryradiomics}, where a large amount of custom, quantitative radiomic features
  are directly discovered from the wealth of readily available medical imaging data.

  In this study, we present a novel
  discovery radiomics framework for skin cancer detection, where we leverage novel multi-column
  deep radiomic sequencers for high-throughput discovery and extraction of a large amount of
  custom radiomic features tailored for characterizing unique skin cancer tissue phenotype.
  Deep neural networks have shown that they can learn effective and accurate feature extraction framework via convolutional layers. This type of operations decrease the human model bias since they are trained as an end-to-end systems. Due to this fact, the features extracted from deep neural networks (particularly convolutional neural networks) have showing promising results in different applications such as object classification~\cite{CNN1,CNN2}, object segmentation and detection~\cite{Seg} and super-resolution~\cite{supres}.

  The proposed framework has considerable potential for discovering a large amount of quantitative biomarkers beyond what clinicians can visually identify, thus making it a
 powerful tool for assisting general practitioners and dermatologists alike in improving the efficiency, consistency, and accuracy of skin cancer diagnosis.

This paper is organized as follows; In the next section the notion of discovery radiomics and the deep multi-column neural network for the purpose of generating discovery radiomic sequences is explained. Then the Results are demonstrated and the conclusion will be drawn at the end.
%-------------------------------------------------------------------------
\section{Methodology}

\begin{figure*}[t]
	\begin{center}
		\includegraphics[width = 14 cm]{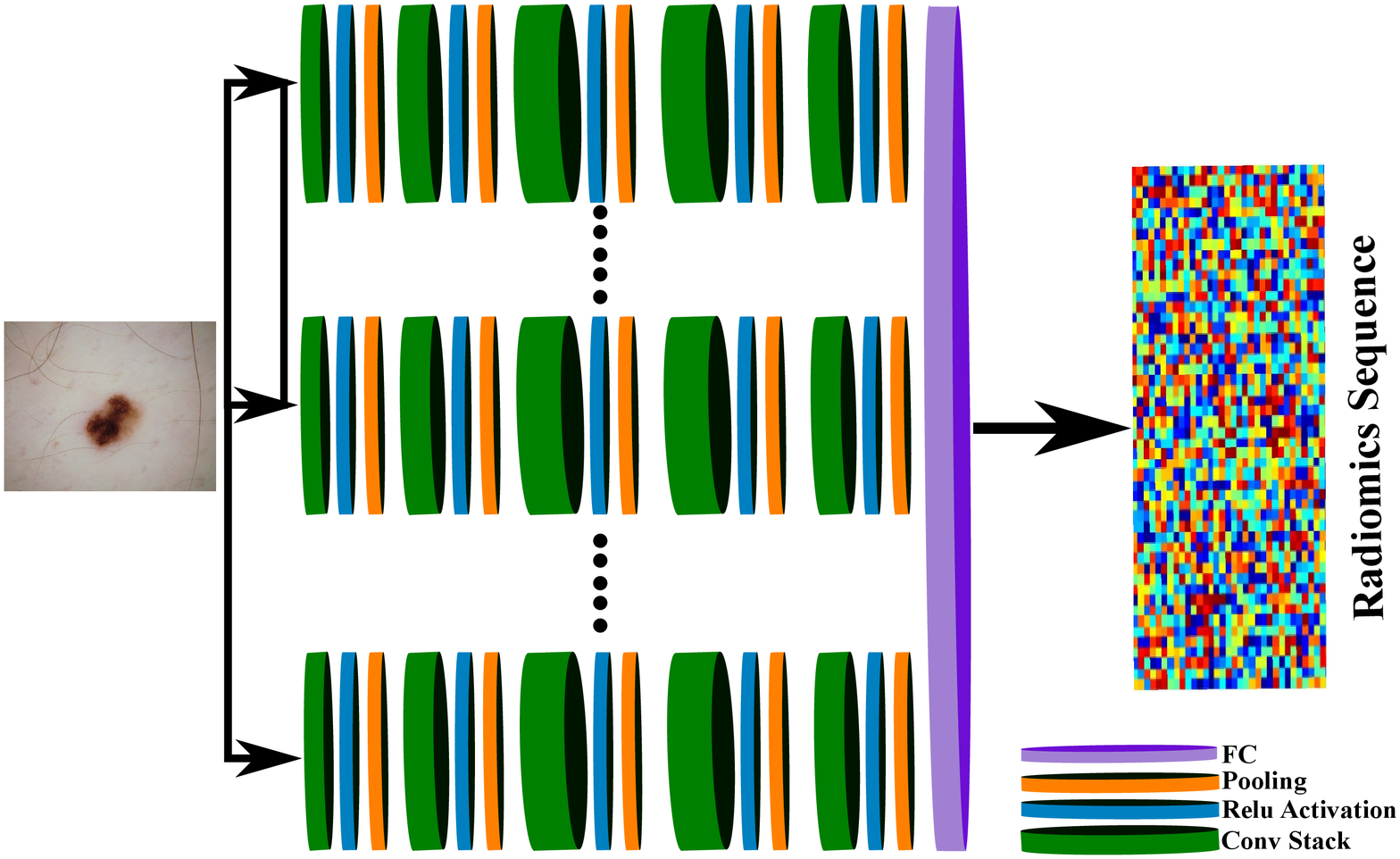}
	\end{center}
	\caption{Overview of the proposed deep multi-column radiomic sequencer. The network architecture of the radiomic sequencer is inspired by~\cite{CNN1}. The network architecture is a multi-column network comprising of two individual columns of convolutional,
		activation, and pooling layers. Each column of the deep multi-column radiomic sequencer is composed of 5 convolutional layers, with the number of convolutional filters are increased and the size of the convolutional filters are decreased as we go deeper to improve the modeling performance of the deep multi-column radiomic sequencer. }
	\label{fig:network}
\end{figure*}

The proposed discovery radiomics framework for skin cancer detection can be described as follows (see Figure~\ref{fig:skinradiomics}). Given
a wealth of past clinical images and corresponding biopsy-verified diagnostic information from a skin
imaging data archive, the radiomic sequencer discovery process discovers a radiomic sequencer
that can perform high-throughput extraction of radiomic sequences comprising of a large amount of highly customized, quantitative features tailored for characterizing unique skin cancer traits that are particularly effective at differentiating between malignant and benign skin lesions. The discovered radiomic sequencer can then be applied to clinical images of a new patient to extract the corresponding dermatological radiomic sequence for skin cancer screening and diagnosis purposes.

In this study, the radiomic sequencer being proposed in the discovery radiomics framework is a novel deep multi-column radiomic sequencer, inspired by the work of Ciresan et al.~\cite{Ciresan}.  More specifically, the proposed deep multi-column radiomic sequencer splits information into multiple, parallel columns so that parallel streams of increasingly more abstract skin cancer traits are discovered and modeled based on a wealth of skin imaging data before being merged into a single representation, thus allowing for improved and more complete characterization of the complex physiological characteristics of skin cancer.

The underlying architecture of the proposed deep multi-column radiomic sequencer being discovered is shown in Figure~\ref{fig:network}.  In this architecture, low-level skin tissue characteristics are modeled in the lower convolutional layers. However the layers are divided into multiple parallel columns of deep convolutional layers that decompose into unique mid- to high-level skin tissue characteristics.  These parallel columns of deep convolutional layers are then merged via a fully-connected layer to produce a final radiomic sequence for the skin lesion being analyzed.

In this study, taking inspiration from~\cite{CNN1},  the realization of the deep multi-column network architecture consists of two columns, with each
column consisting of five convolutional layers. The examined network architecture in this study is the combination of two-column networks with each column consisting of 5 convolutional layers.  The number of convolutional filters are increased and the size of the convolutional filters are decreased as we go deeper to improve the modeling performance of the deep, multi-column radiomic sequencer.  This led the proposed radiomic sequencer to possess the following architectural configuration for each column: five convolutional layers with 96@ $7\times7$ filters, 256@ $5\times5$ filters, 384@ $3\times3$ filters, 384@ $3\times3$ filters, 256@ $5\times5$ filters, and one fully connected layer with 8192 neurons. The output of two columns are combined together to produce the final radiomic sequence.

\subsection{Results and Discussion}

\begin{figure}
	\begin{center}
		\includegraphics[width = 9 cm]{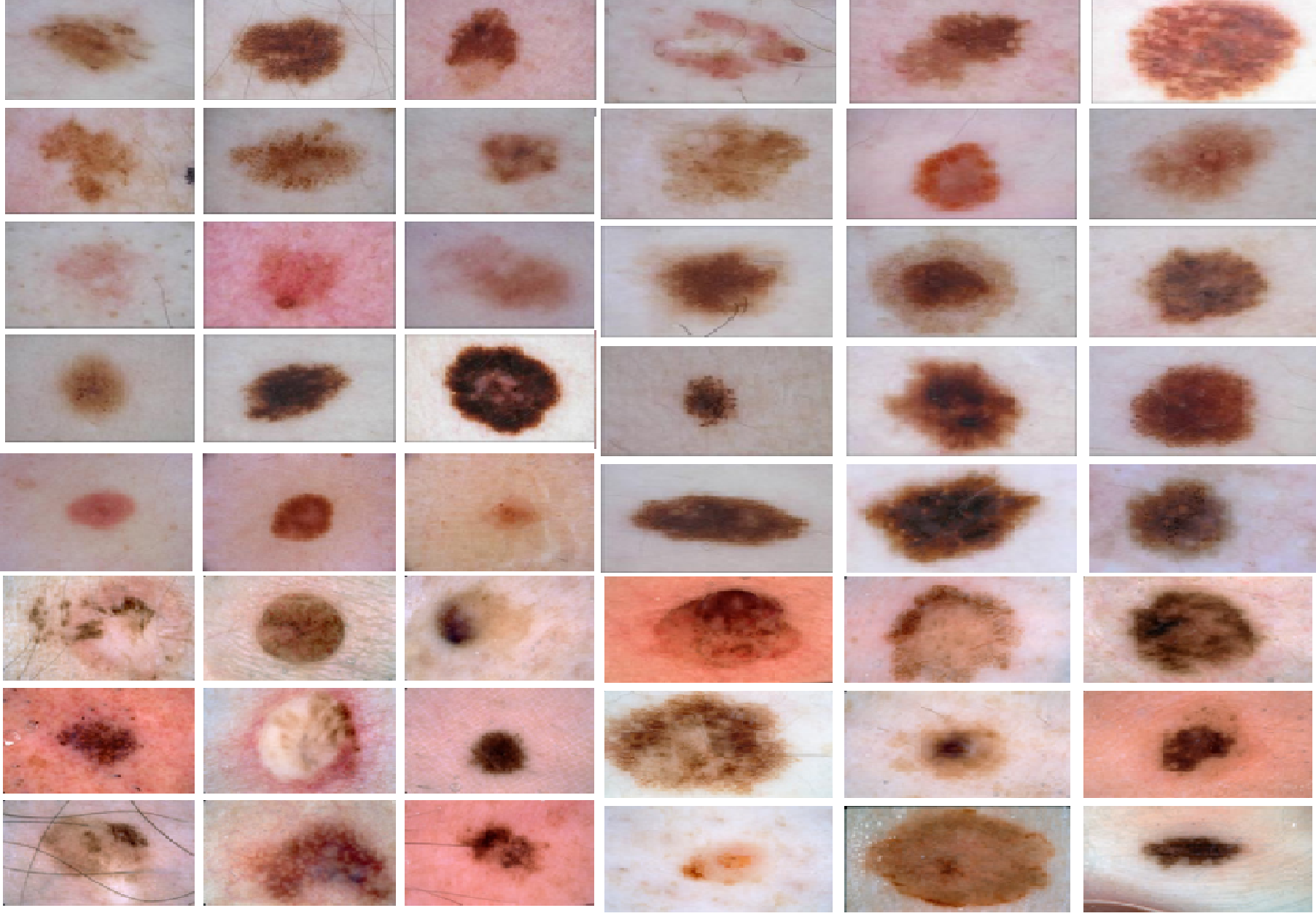}
	\end{center}
	\caption{Examples of biopsy-proven clinical images of skin lesions from~\cite{ISIC-Skin}}
	\label{fig:ISIC}
\end{figure}

  In this study, the discovered deep multi-column radiomic sequencer was tested against 9,152 biopsy-proven clinical images extracted from ISIC 2017 Challenge dataset~\cite{ISIC-Skin} (see Figure~\ref{fig:ISIC}), with the malignant cases comprising of 90\% melanoma cases and 10\% basal cell carcinoma cases.  Sequencer validation was performed by splitting the clinical images into 10\% training (473 randomly selected samples for each class label) and 90\% testing. The testing dataset comprises of 8,049 benign and 157 malignant cases.    To quantitatively evaluate the efficacy of the discovery radiomic sequencer for cancer detection, the radiomic sequences produced using the sequencer are then fed into a fully-connected feed-foward neural network with two fully-connected layers and a softmax layer.

   The efficacy of the proposed discovery radiomics framework is examined quantitatively using two performance metrics: i) sensitivity, and ii) specificity; where
   \begin{align}
   \text{sensitivity} = \frac{\text{true positive}}{\text{positive}}
   \end{align}
\begin{align}
\text{specificity} = \frac{\text{true negative}}{\text{negative}}.
\end{align}
where true positive is the number of malignant cases which are classified correctly by the proposed approach and true negative is the number of benign cases classified as negative by the proposed approach.

   Experimental results show that the proposed radiomic sequencer achieved sensitivity and specificity of 91\% and 75\%, respectively.  As a point of reference, in the study by Wells et al.~\cite{Wells}, it was found that the dermatologists in the study had a sensitivity and specificity of 80\% and 43\%, respectively for melanoma screening, while MelaFind, a non-invasive light-based tool for skin cancer screening, had a sensitivity and specificity of 96\% and 8\%, respectively.  Furthermore, in a study by Esteva et al.~\cite{Esteva}, it was found that using an Inception v3 convolutional neural network trained for distinguishing between benign skin lesions, malignant lesions, and non-neoplastic lesions achieved an accuracy of 72.1 $\pm$ 0.9\% on dermatologist-labeled clinical images.  As such, these experimental results show that the proposed radiomic sequencer for skin cancer detection is able to achieve dermatologist-level
  performance and hence can be a powerful tool for assisting general practitioners and dermatologists
  alike in improving the efficiency, consistency, and accuracy of skin cancer diagnosis.

\section{Conclusion}
In this paper we proposed a new discovery radiomics approach for the purpose of skin cancer modeling and classification. A deep multi-column radiomic sequencer is proposed and discovered for high-throughput discovery and extraction of a large amount of
  custom radiomic features tailored for characterizing unique skin cancer tissue phenotype.  The deep multi-column architecture used in the proposed radiomic sequencer can significantly boost the modeling power of the sequencer.  The discovered deep multi-column radiomic sequencer can then be applied to clinical images of a new patient to extract the corresponding dermatological radiomic sequence for skin cancer screening and diagnosis purposes.

The proposed framework is examined on ISIC skin cancer challenge, with the deep multi-column radiomic sequencer discovered using a small balanced dataset of malignant and benign cases where only 473 cases were utilized for each class label. Quantitative evaluation of the discovered radiomic sequencer was then performed using an unbalanced dataset with 8,049 benign cases and 157 malignant cases. Experimental results demonstrated that the proposed discovery radiomics approach for skin cancer modeling and classification is able to achieving sensitivity and specificity of 91\% and 75\%, respectively.  These promising results show the applicability and modeling power of the proposed approach and illustrates that it can be a powerful tool for assisting general practitioners and dermatologists alike in improving the efficiency, consistency, and accuracy of skin cancer diagnosis.

\section*{Acknowledgments}
This work was supported by Elucid Labs. The authors also thank Nvidia for the GPU hardware used in this study through the Nvidia Hardware Grant Program.


\begin{thebibliography}{99}
	\bibitem{CNN1} A. Krizhevsky,  I. Sutskever, and G. Hinton. Imagenet classification with deep convolutional neural networks \emph{Advances in neural information processing systems (NIPS)}(2012).
	\bibitem{CNN2}  K. Simonyan, and A. Zisserman. Very deep convolutional networks for large-scale image recognition \emph{arXiv preprint arXiv:1409.1556}(2014).
\bibitem{Seg} J. Long,  E. Shelhamer,  and T. Darrell. Fully convolutional networks for semantic segmentation \emph{IEEE Conference on Computer Vision and Pattern Recognition (CVPR)} (2015).

\bibitem{supres} C. Dong,  C. Loy, K. He and X. Tang.  Learning a deep convolutional network for image super-resolution \emph{European Conference on Computer Vision}(2014).
\bibitem{Esteva}
 A.~Esteva, and B.~Kuprel
\newblock Dermatologist-level classification of skin cancer with deep neural
networks. \newblock {\em Nature}, (2016).


\bibitem{Ciresan}
D.~Ciresan, U.~Meier, and J.~Schmidhuber.
\newblock Multicolumn deep neural networks for image classification.
\newblock {\em IEEE Conference on Computer Vision and Pattern Recognition},(2012).


\bibitem{Guy}
G.~GP, M.~SR, E.~DU, and Y.~KR.
\newblock Prevalence and costs of skin cancer treatment in the u.s., 2002-2006 and 2007-2011.
\newblock {\em Am J Prev Med}, 2014.


\bibitem{discoveryradiomics2}
A.-H. Karimi, A.~G. Chung, M.~J. Shafiee, F.~Khalvati, M.~A. Haider, A.~Ghodsi,and A.~Wong.
\newblock Discovery radiomics via a mixture of deep convnet sequencers for
multi-parametric mri prostate cancer classification.
\newblock {\em International Conference Image Analysis and Recognition}, (2017).


\bibitem{discoveryradiomics3}
D.~Kumar, A.~G. Chung, M.~J. Shafiee, F.~Khalvati, M.~A. Haider, and A.~Wong.
\newblock Discovery radiomics for pathologically-proven computed tomography
lung cancer prediction.
\newblock {\em International Conference Image Analysis and Recognition}, 2017.


\bibitem{radiomics}
P.~Lambin, E.~Rios-Velazquez, R.~Leijenaar, S.~Carvalho, R.~G. van Stiphout,
P.~Granton, C.~M. Zegers, R.~Gillies, R.~Boellard, and A.~D. et~al.
\newblock Radiomics: extracting more information from medical images using
advanced feature analysis.
\newblock {\em European Journal of Cancer}, 2012.

\bibitem{radiomics2}
A.~Cameron, F.~Khalvati, M.~Haider, and A.~Wong.
\newblock MAPS: A Quantitative Radiomics Approach for Prostate Cancer Detection.
\newblock {\em IEEE Transactions on Biomedical Engineering}, (2016).

\bibitem{radiomics3}
R.~Amelard, J.~Glaister, A.~Wong, and D.~Clausi.
\newblock High-level intuitive features (HLIFs) for intuitive skin lesion description.
\newblock {\em IEEE Transactions on Biomedical Engineering}, (2015).

\bibitem{discoveryradiomics1}
M.~J. Shafiee, A.~G. Chung, D.~Kumar, F.~Khalvati, M.~Haider, and A.~Wong.
\newblock Discovery radiomics via stochasticnet sequencers for cancer
detection. \newblock {\em NIPS Workshop on Machine Learning in Healthcare}, (2015).


\bibitem{Stern}
R.~Stern.
\newblock Prevalence of a history of skin cancer in 2007: results of an
incidence-based model.
\newblock {\em Arch Dermatol}, (2010).


\bibitem{Wells}
R.~Wells, D.~Gutkowicz-Krusin, and e.~a. E.~Veledar.
\newblock Comparison of diagnostic and management sensitivity to melanoma
between dermatologists and melafind: A pilot study.
\newblock {\em International Conference Image Analysis and Recognition}, (2012).


\bibitem{discoveryradiomics}
A.~Wong, A.~G. Chung, D.~Kumar, M.~J. Shafiee, F.~Khalvati, and M.~Haider.
\newblock Discovery radiomics for imaging-driven quantitative personalized
cancer decision support.
\newblock {\em Journal of Comptutational Vision and Imaging Systems}, (2015).
	
\bibitem{ISIC-Skin} D. Gutman,  N.  Codella, E.  Celebi, B. Helba, M. Marchetti,N.  Mishra, Nabin and A.  Halpern. Skin lesion analysis toward melanoma detection: A challenge at the international symposium on biomedical imaging (ISBI) 2016, hosted by the international skin imaging collaboration (ISIC)\emph{arXiv preprint arXiv:1605.01397}(2016).

	
\end{thebibliography}
\end{document}